# Multi-Head Spectral-Adaptive Graph Anomaly Detection

Cao Qing Yue
People's Public Security University of China
Beijing, China
qingyue-c@outlook.com

Jin Bo
Third Research Institute of the Ministry of Public Security
Shanghai, China
jinbo@gass.cn

Gong Chang Wei
People's Public Security University of China
Beijing, China
2025111035@stu.ppsuc.edu.cn

Tong Xin
People's Public Security University of China
Beijing, China
tongxindotnet@outlook.com

Li Wen Zheng
People's Public Security University of China
Beijing, China
sdrzlwz@126.com

Zhou Xiao Dong
Shanghai Police College
Shanghai, China
zhouxdshpc@163.com

## ABSTRACT

Graph anomaly detection technology has broad applications in financial fraud and risk control. However, existing graph anomaly detection methods often face significant challenges when dealing with complex and variable abnormal patterns, as anomalous nodes are often disguised and mixed with normal nodes, leading to the coexistence of homophily and heterophily in the graph domain. Recent spectral graph neural networks have made notable progress in addressing this issue; however, current techniques typically employ fixed, globally shared filters. This 'one-size-fits-all' approach can easily cause over-smoothing, erasing critical high-frequency signals needed for fraud detection, and lacks adaptive capabilities for different graph instances. To solve this problem, we propose a Multi-Head Spectral-Adaptive Graph Neural Network (MHSA-GNN). The core innovation is the design of a lightweight hypernetwork that, conditioned on a 'spectral fingerprint' containing structural statistics and Rayleigh quotient features, dynamically generates Chebyshev filter parameters tailored to each instance. This enables a customized filtering strategy for each node and its local subgraph. Additionally, to prevent mode collapse in the multi-head mechanism, we introduce a novel dual regularization strategy that combines teacher-student contrastive learning (TSC) to ensure representation accuracy and Barlow Twins diversity loss (BTD) to enforce orthogonality among heads. Extensive experiments on four real-world datasets demonstrate that our method effectively preserves high-frequency abnormal signals and significantly outperforms existing state-of-the-art methods, especially showing excellent robustness on highly heterogeneous datasets.

## 1 INTRODUCTION

Financial fraud detection has evolved from identifying isolated outliers to uncovering complex, camouflaged subgraphs within massive transaction networks [1]. Graph Neural Networks (GNNs) have emerged as a key technology for capturing latent fraudulent patterns in transaction networks, owing to their exceptional capability in modeling non-Euclidean data [2-4]. While applying them to financial risk control remains challenging, the core conflict lies in the design of classical GNNs (e.g., GCN [5], GraphSAGE [6]), which inherently function as low-pass filters [7]. These models rely on the homophily assumption to smooth feature signals, which inadvertently leads to "over-smoothing" in fraud scenarios where fraudsters actively establish high-heterophily connections with normal users to camouflage themselves [5, 8, 9].

To overcome the limitations of spatial aggregation, researchers have turned to the spectral domain, where anomalies typically manifest as high-frequency signals or cause a "right-shift" in spectral energy distribution [10, 11]. Consequently, the frontier of fraud detection has shifted towards designing filters capable of capturing full-spectrum information [12, 13]. Early attempts using polynomial approximations (e.g., GIN [14], ChebNet [15], DeepWalk [16]) theoretically offer universal approximation but struggle with high-order instability in large financial graphs. Subsequent studies have demonstrated that while standard models like GAT [17] fail to capture high-frequency components, explicitly designed spectral models (e.g., CayleyNet [18]) can mitigate this issue [2, 19, 20]. However, existing spectral GNNs still face significant limitations when dealing with the extreme heterogeneity and adversarial nature of financial fraud.

The first major limitation is the reliance on globally shared filters. Most advanced methods, such as BernNet [12], AutoGCN [21], and various polynomial-based approaches, optimize a single set of filter parameters for the entire graph. While recent works like BWGNN [24] and DSGAD [26] introduce Beta wavelets to handle band-pass filtering, and others explore frequency localization or structural roles [25, 27, 28], they still largely depend on predefined bases or global optimization. This "one-size-fits-all" paradigm fails to account for the instance-level diversity of fraud patterns (e.g., money laundering rings vs. credit card theft), where local spectral fingerprints vary dramatically. Other attempts to handle heterophily via edge pruning (GHRN [22]) or node-level homophily estimation (LH-GNN [23]) often risk discarding critical fraud indicators or suffer from label noise.

The second limitation lies in the redundancy and lack of adaptivity in multi-frequency architectures. To capture diverse patterns, recent SOTA methods have adopted complex multi-channel or multi-head designs [29-34]. For instance, NMFA [29] and GraphPN [30] utilize multi-head attention or band control, while ChiGAD [31], SComGNN [32], and AdaGNN [34] design specific filters for heterogeneous graphs or different frequency bands. However, without effective regularization, these multi-head mechanisms often suffer from mode collapse, converging to redundant low-frequency responses [33]. Furthermore, methods relying on fixed bases or complex decompositions (e.g., Wiener filters or meta-paths) incur high computational costs, making them unsuitable for real-time financial systems.

To bridge these gaps, we propose MHSA-GNN, a Multi-Head Spectral-Adaptive Graph Neural Network driven by spectral fingerprints. Unlike previous methods that learn static parameters, MHSA-GNN introduces a HyperNetwork mechanism that dynamically generates instance-specific Chebyshev filter coefficients based on a compact spectral fingerprint of the local subgraph, which achieves true instance-level adaptation. Furthermore, we design a dual regularization strategy combining Teacher-Student Contrastive Learning and Barlow Twins Diversity, mathematically forcing different heads to learn orthogonal spectral features.

Our main contributions are summarized as follows:

**Instance-Level Spectral Adaptation**: We propose a lightweight spectral fingerprinting mechanism that enables the model to generate "tailor-made" filter parameters for each subgraph, breaking the bottleneck of global shared filters.

**Orthogonal Multi-Head Learning**: We introduce a dual regularization scheme that prevents mode collapse by enforcing statistical orthogonality among attention heads, implicitly achieving automatic frequency band decomposition.

**Superior Performance**: Extensive experiments demonstrate that MHSA-GNN outperforms SOTA spectral GNNs in detecting camouflaged fraud patterns.

## 2  Preliminary

Formally, we define an attributed graph as $G = (V, E, X)$, where $V$ is the set of N nodes, $E$ represents the set of edges, and $X \in \mathbb{R}^{N \times F}$ denotes the node feature matrix. The graph structure is characterized by the normalized Laplacian operator $L = I - D^{-\frac{1}{2}} A D^{-\frac{1}{2}}$, with its eigendecomposition given by $L = U \Lambda U^T$, Here, $U$ forms the basis of the Graph Fourier Transform, $U^T x$ transforms the node signal x into the spectral domain, A is the adjacency matrix, D is the degree matrix, and I is the identity matrix.

In the context of graph signal processing, a graph convolution is intrinsically a graph filter, where the filtering operation on node features x is expressed as the product with a spectral filter $g_\theta(\cdot)$. In the spectral domain, a graph filter $g(\lambda)$ is a function defined on the eigenvalues $\lambda$ of the graph Laplacian L. While prior works such as ChebyNet approximate this function using a K-th order polynomial $g_\theta(\Lambda) \approx \sum_{k=0}^{K} \theta_k \lambda^k$, to circumvent the computationally prohibitive eigendecomposition and enable efficient computation in the spatial domain, K-th order Chebyshev polynomials $T_k(\cdot)$ are typically employed to approximate $g_\theta(\tilde{L})$:

$$Z = g_\theta(\tilde{L}) * X \approx \sum_{k=0}^{K} \theta_k T_k(\tilde{L}) * X \quad (1)$$

Here, $Z \in \mathbb{R}^{N \times F'}$ represents the filtered node representations, and $\theta = [\theta_0, \theta_1, \ldots, \theta_K]$ is the polynomial coefficient vector that uniquely determines the frequency response characteristics of the filter (including low-pass, band-pass, or high-pass properties). $\tilde{L} = \frac{2}{\lambda_{max}} L - I$ denotes the rescaled Laplacian mapping eigenvalues to the interval [-1,1], where $\lambda_{max}$ is the largest eigenvalue of L. The term $T_k(\tilde{L})$ refers to the Chebyshev polynomial of order k, computed recursively via $T_k(x) = 2x T_{k-1}(x) - T_{k-2}(x)$.

## 3  METHODOLOGY

We propose a Multi-Head Spectral-Adaptive Graph Neural Network tailored for graph anomaly detection tasks. In contrast to traditional GNNs (e.g., GCN or ChebyNet) that utilize fixed and globally shared filters (typically fixed low-pass filters), the core of our proposed dynamic spectral-adaptive model lies not in learning a static set of polynomial coefficients, but in dynamically generating filter coefficients optimal for the specific graph via a generation network conditioned on the input graph's spectral fingerprint. Furthermore, the model operates multiple independent spectral-adaptive filter heads in parallel and employs a regularization strategy to prevent multi-head mode collapse (i.e., all heads learning identical representations), thereby promoting functional specialization among the filter heads and enhancing the model's capacity to capture diverse anomaly patterns.

### 3.1  Spectral-Adaptive Filter Parameter Generation

Conventional approaches treat the coefficients θ as globally shared, learnable parameters, implying that the model learns only an "average optimal" filter (e.g., forced low-pass in GCN), which neglects the significant variations in topological structure and signal distribution across different graphs or regions. We assume that the optimal filter should vary in accordance with the graph data. To this end, we design a parameter generation network $\phi(\cdot)$ capable of dynamically generating the most suitable coefficients based on the current graph's "fingerprint".

spectral fingerprint: We construct a compact 20-dimensional vector $f_{spec} \in \mathbb{R}^{20}$ to comprehensively characterize the graph's spectral properties, consisting of two components:

(1)**Structural Spectral Fingerprint**( $f_{struct} \in \mathbb{R}^4$ ): This describes the macroscopic topological structure of the graph. It is constructed by calculating the $w$(we set $w$=6) largest and smallest eigenvalues $\lambda_i$ of the graph Laplacian L and extracting statistical moments of these eigenvalues. The value of $w$ will be adjusted according to the graph size to ensure they do not exceed the number of eigenvalues allowed by the matrix:

$$f_{struct} = [mean(\lambda), var(\lambda), skew(\lambda), kurtosis(\lambda)] \quad (2)$$

The mean and variance describe the central tendency and dispersion of the graph spectrum, respectively. Skewness measures the asymmetry, and kurtosis measures the peakedness or flatness of the distribution. Together, these statistical features describe the frequency distribution pattern of the graph. To ensure scalability to large-scale graphs, we employ stochastic spectral estimation, utilizing the Hutchinson trace estimator and the stochastic Lanczos method [35] to efficiently approximate these moments, avoiding expensive full matrix eigendecomposition.

(2)**Signal Fingerprint**($f_{signal} \epsilon \mathbb{R}^{16}$): This characterizes the smoothness of node features on the graph, measured by the Rayleigh quotient of the feature signal X on the graph L:

$$\rho(L,X) = \frac{x^T L x}{x^T x} \quad (3)$$

A smaller value of $\rho(L,X)$ indicates a smoother signal x (values of X at adjacent nodes are close), representing a low-frequency signal; conversely, $\rho(L,X)$ approaches $\lambda_{max}$, it implies signal oscillation, indicating a high-frequency signal. We project the raw features X into a 16-dimensional space using a fixed random matrix to obtain $X_{proj}$ and subsequently compute the Rayleigh quotient for each column (feature dimension $x_i$) of $X_{proj}$ to construct $f_{signal}$.

$$\rho(x_i) = \frac{x_i^T L x_i}{x_i^T x_i} \quad (4)$$

The final spectral fingerprint is the concatenation of these two components:

$$f_{spec}(G,X) = Concat(f_{struct}, f_{signal}) \quad (5)$$

Subsequently, we utilize a small neural network $\phi$ (a two-layer MLP) to dynamically generate the Chebyshev coefficients $\theta$, where K is the polynomial order:

$$\theta = \{\theta_0, \theta_1, ..., \theta_K\} = \phi(f_{spec}) \quad (6)$$

Finally, using this generated set of $\theta$, we perform standard Chebyshev polynomial convolution, followed by a linear transformation and activation function to obtain the final node representation:

$$Z = LeakyReLU(W \cdot \sum_{k=0}^{K} \theta_k T_k(\tilde{L}) * X + B) \quad (7)$$

Conceptually, when the input graph's $f_{signal}$ indicates high-frequency energy, $\phi$ outputs parameters $\theta$ constituting a high-pass filter; conversely, when $f_{struct}$ indicates energy concentration in low frequencies, $\phi$ generates parameters for a low-pass filter.

### 3.2 Multi-Head Frequency Analysis

To capture complex anomaly patterns from multiple frequency perspectives, we extend the aforementioned single spectral-adaptive filter into a multi-head architecture, specifically a parallel spectral filter composed of H spectral-adaptive convolution heads. The parameter generation network (Eq. 6) is extended to accept a single spectral fingerprint $f_{spec}$ as input and generate a unique set of Chebyshev coefficients $\theta^{(h)}$ for each head $h \epsilon \{1, ..., H\}$:

$$\{\theta^{(1)}, \theta^{(2)}, ..., \theta^{(H)}\} = \phi_{multi-head}(f_{spec}) \quad (8)$$

Here, $\theta^{(h)} = \{\theta_0^{(h)}, \theta_1^{(h)}, ..., \theta_K^{(h)}\}$. Each filter head h subsequently uses its exclusive coefficients $\theta^{(h)}$ to perform parallel spectral filtering (Eq. 7) on the input features X, yielding H distinct sets of node representations $\{Z^{(1)}, Z^{(2)}, ..., Z^{(H)}\}$.

To adaptively aggregate these diverse representations, we employ a channel attention mechanism. For each node i, we first calculate the importance weight $\alpha_i^{(h)}$ of its representation $z_i^{(h)}$ in each head h, and then compute the weighted sum to obtain the final fused representation $z_i^{fuse}$:

$$\alpha_i^{(h)} = softmax_h(MLP_{attn}(z_i^{(h)})) \quad (9)$$

$$z_i^{fuse} = \sum_{h=1}^{H} \alpha_i^{(h)} z_i^{(h)} \quad (10)$$

where $\alpha_i^{(h)}$ denotes the weight of node i with respect to filter head h.

### 3.3 Regularization for Multi-Filter Specialization

A common issue in learnable filter banks is mode collapse: without additional constraints, given that graph signals typically exhibit the strongest energy in the low-frequency band, the H filters are prone to converging to the same optimal low-pass filter coefficients (i.e., $\theta^{(1)} \approx \theta^{(2)} \approx \cdots \approx \theta^{(H)}$), resulting in severe information redundancy. To prevent the learning of redundant information and enforce head specialization, we design a robust dual self-supervised regularization strategy combining Filter Trajectory Stability Constraint and Spectral Response Decorrelation. The former enforces each filter to learn meaningful representations consistent with a global view via a Teacher-Student architecture; the latter mandates orthogonality among representations from different heads, compelling them to capture distinct frequency bands of the graph signal, thereby guiding the model to learn diverse and complementary representations.

(1)**Teacher-Student Based Filter Trajectory Stability Constraint:** Designed to enhance the uniqueness of each head's representation. We maintain two models with identical architectures:

**Student Network S**: parameterized by $\xi_S$, updated normally via gradient descent during training.

**Teacher Network T**: acts as a temporal smoother for filter evolution, parameterized by $\xi_T$, and $\xi_T$ is not updated via backpropagation but slowly updated as an Exponential Moving Average (EMA) [36] of the Student parameter:

$$\xi_T \leftarrow m\xi_T + (1-m)\xi_S \quad (11)$$

where m is the momentum coefficient. Network T, due to its smooth parameter updates, provides a more stable set of target representations. During training, Network S outputs parallel filter bank representations $\{z_1^S, ..., z_H^S\}$, and Network T similarly outputs H stable filter representations $\{z_1^T, ..., z_H^T\}$.

We employ the InfoNCE contrastive loss to train Network S. For the i-th filter output representation $z_i^S$ of the Student network, the corresponding Teacher filter output $z_i^T$ is treated as the positive sample, while all other Teacher filter representations $z_j^T$ (j≠ i) are treated as negative samples. This forces $z_i^S$ to match the unique signal of $z_i^T$ rather than the signals of $z_j^T$:

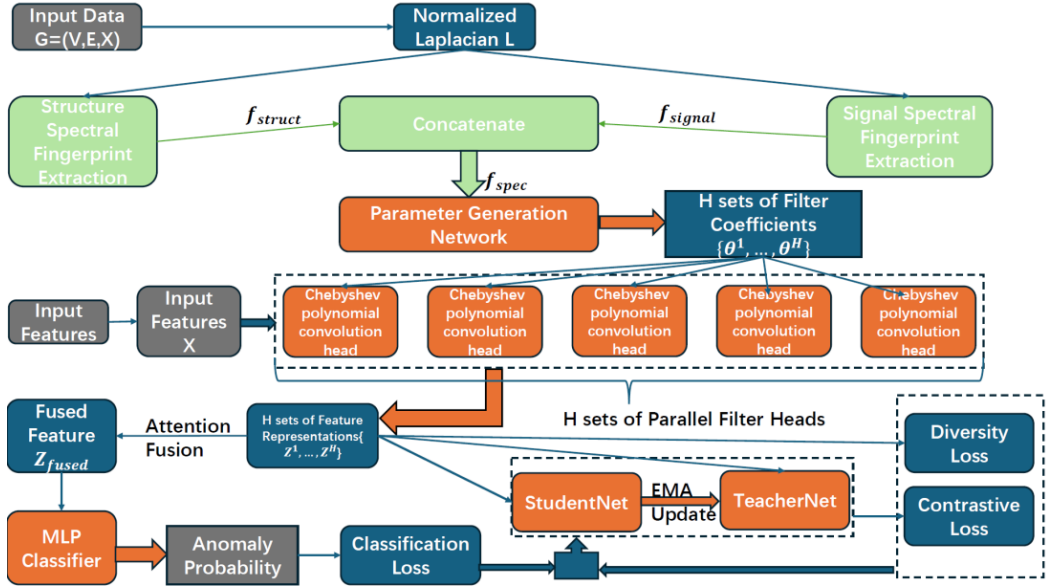

**Figure 1. Workflow of MHSA-GNN**

$$\mathcal{L}_{contrast} = -\sum_{i=1}^{H} \log \frac{\exp(cosine\_sim(z_i^S, z_i^T)/\tau)}{\sum_{j=1}^{H} \exp(cosine\_sim(z_i^S, z_j^T)/\tau)} \quad (12)$$

where cosine_sim computes cosine similarity and $\tau$ is a temperature hyperparameter. $\mathcal{L}_{contrast}$ serves as a constraint for smooth and stable filter parameter trajectories, enforcing the H filters to track unique targets respectively, thereby achieving functional separation.

(2)**Barlow Twins Based Decorrelation**: To further eliminate information redundancy across heads, we introduce the Barlow Twins loss [37] to enforce feature orthogonality. First, the outputs of the H Student filters (after normalization and projection) are concatenated along the feature dimension to obtain a batch representation $Z_{batch}$. Subsequently, we compute the cross-correlation matrix $C \in \mathbb{R}^{D' \times D'}$ of the feature dimensions of $Z_{batch}$ (where $D'$ is the total feature dimension after projection). This mechanism enforces mutual orthogonality among representations from different heads. The loss function minimizes the difference between $C$ and the identity matrix I:

$$\mathcal{L}_{diversity} = \sum_{i,j}(C_{ij} - I_{ij})^2 = \sum_{i=1}^{D'}(C_{ii} - 1)^2 + \sum_{i \neq j} C_{ij}^2 \quad (13)$$

Minimizing the cross-filter decorrelation loss $\mathcal{L}_{diversity}$ encourages $C_{ii} \to 1$ to preserve the information within each head, and $C_{ij} \to 0$ to eliminate information redundancy between filters, removing correlations between different feature dimensions. Since different $z^{(h)}$ contribute to distinct feature components in $Z_{batch}$, this is equivalent to forcing the feature representations extracted by different filters to be statistically orthogonal, effectively achieving specialization and independence.

### 3.4 Training Objective and Optimization

The algorithmic workflow of our proposed MHSA-GNN is illustrated in Figure 1. The model training is performed via end-to-end optimization using a composite loss function:

$$\mathcal{L}_{total} = \mathcal{L}_{class} + \lambda_{contrast} \cdot \mathcal{L}_{contrast} + \lambda_{div} \cdot \mathcal{L}_{diversity} \quad (14)$$

Here, $\mathcal{L}_{class}$ denotes the weighted cross-entropy loss, utilized to address the class imbalance inherent in graph anomaly detection tasks. $\lambda_{contrast}$ and $\lambda_{div}$ are hyperparameters balancing the contribution of each loss component, while $\mathcal{L}_{contrast}$ and $\mathcal{L}_{diversity}$ serve as regularization terms.

Direct optimization of Equation 14 can be unstable, primarily because, in the early stages of training, a randomly initialized Teacher network fails to provide meaningful contrastive targets. To mitigate this, we introduce a critical Teacher Warm-up Mechanism:

**Warm-up Phase**: $\lambda_{contrast}$ is set to 0. The model is trained exclusively using $\mathcal{L}_{class}$ and $\mathcal{L}_{diversity}$, allowing the Student network to learn preliminary stable representations.

**Formal Training Phase**: $\lambda_{contrast}$ is activated. At this stage, the Student network has stabilized, and the Teacher (evolving as the EMA of the Student) is capable of providing meaningful and smooth supervisory signals, allowing $\mathcal{L}_{contrast}$ to guide the specialization of the filters.

### 3.5 Analysis of Spectral-Adaptive Filter Parameters

To validate the proposed multi-head spectral-adaptive network—specifically its ability to circumvent the limitations of global fixed filters in traditional GNNs and achieve adaptive filter design for distinct graph instances—we conducted a series of experiments. These experiments elucidate how the hypernetwork dynamically generates filter parameters with varying frequency response characteristics for normal versus fraudulent nodes based on input spectral fingerprints, thereby verifying the model's discriminative capability in handling low-frequency homophilic signals and high-frequency heterophilic signals.

We performed these experiments on two representative datasets: Amazon (characterized by prominent structural features) and T-finance (characterized by complex structure and high camouflage):

**Node Sampling and Subgraph Extraction:** We randomly sampled pairs of normal and fraudulent nodes from the test set and extracted the 2-hop subgraph for each node. The selection of 2-hop subgraphs aligns with the receptive field of the Chebyshev polynomial convolution, ensuring the capture of complete spectral information characterizing the local node topology.

**Spectral Fingerprint Calculation and Parameter Generation:** For each subgraph, we computed the 20-dimensional spectral fingerprint (comprising structural statistics and signal smoothness) and fed it into the trained parameter generation network to extract the output multi-head Chebyshev coefficients.

**Frequency Domain Transformation and Visualization:** Leveraging the properties of Chebyshev polynomials, we transformed the generated coefficients into frequency response functions $g(\lambda)$, where $\lambda$ represents the normalized frequency. We subsequently compared the gain differences between node classes across low and high-frequency regions

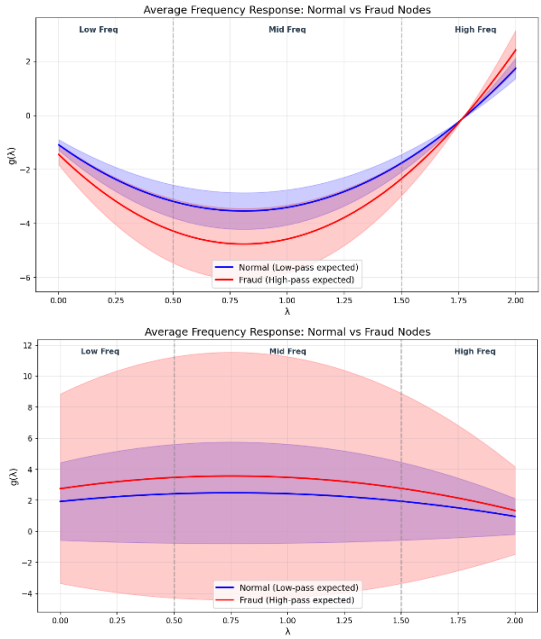

**Figure 2: Analysis of Subgraph Frequency Band Response**

As illustrated in the top graph of Figure 2, the horizontal axis represents the normalized frequency $\lambda$ (ranging from 0 for low frequency to 2 for high frequency), while the vertical axis $g(\lambda)$ indicates the filter gain or response intensity.

The blue curve represents the average filter response for normal nodes. It exhibits high intensity at $\lambda=0$ with low variance in average response, consistent with the low-pass filtering mechanism of GCNs that smooths neighbor features.

The red curve represents the average filter response for fraudulent nodes. It shows a sharp rise in the high-frequency region where $\lambda>1.5$. This demonstrates that the model has learned to mine fraudsters hidden among normal nodes by amplifying high-frequency signals.

Conversely, the bottom graph in Figure 2, the trends for normal and fraudulent nodes are highly similar, both exhibiting weak all-pass characteristics. This reflects the low topological distinguishability between normal and fraudulent nodes in the T-finance dataset.

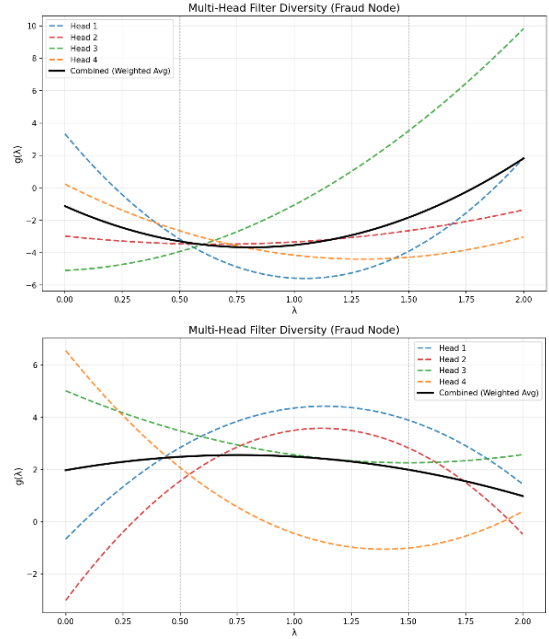

**Figure 3: Analysis of Filter Bank Diversity**

The top graph of Figure 3 displays the shapes of independent filters corresponding to 4 attention heads for a specific fraud node in the Amazon dataset, along with their weighted final effect:

The green dashed line depicts a powerful high-pass filter, and the orange dashed line functions as a low-pass filter, providing smoothing and aggregation capabilities. The blue dashed line represents a band-stop filter, and the red dashed line is an inverted low-pass filter, utilized for contrastive learning within this filter group. The black solid line shows the final weighted result, exhibiting distinct band-stop characteristics with high-frequency enhancement and mid-frequency attenuation. This confirms that our dual regularization strategy successfully prevents mode collapse and achieves true multi-view frequency domain analysis.

The bottom graph of Figure 3 showcases the generation of multi-head spectral-adaptive filters on the T-finance dataset:

The blue and red dashed lines are both mid-frequency band-pass filters, selectively passing mid-to-high frequency information. And the orange and green dashed lines are both low-pass filters; specifically, the green line performs neighborhood aggregation, while the orange line acts as a strong low-pass filter to extract long-term trends in the graph. The black solid line exhibits a "fused low-pass" characteristic, reflecting a strategy dominated by neighborhood aggregation supplemented by mid-to-high frequency information.

This set of experiments demonstrates the model's capability for instance-level adaptive frequency perception.

## 4 EXPERIMENTS

## 4.1 Datasets and Metrics

We evaluate our proposed framework on four benchmark datasets: Amazon [38], T-finance [11], Tolokers [39], and Elliptic [40], as detailed in Table 1. T-Finance is utilized for detecting anomalous accounts in financial transaction networks, and Tolokers is based on worker profile information and task performance statistics, used to predict which workers are banned from specific projects. Amazon is a classic heterogeneous graph dataset in recommendation systems, recording user reviews of products on the Amazon website, where Elliptic is derived from a real-world Bitcoin transaction network, used to detect whether node categories are illicit.

This study employs the two most widely used metrics: AUC and F1-macro. AUC represents the Area Under the ROC Curve and serves as a standard statistic for evaluating classifier predictive capability. F1-macro computes the F1-score independently for each class and then calculates their unweighted arithmetic mean, thereby achieving a harmonic balance between Precision and Recall. Graph anomalies often rely on high-frequency, fragile edge connections; consequently, even minor structural perturbations can disrupt these critical heterogeneous patterns, leading to label semantic drift. In contrast to methods like GraphCL [41] that rely on structural augmentation to construct contrastive views, the output representation diversity in our model is endogenously driven entirely by the initialization differences of the spectral filter banks and the orthogonalization constraints of the Barlow Twins loss. Therefore, MHSA-GNN applies strictly consistent topological inputs to both the Teacher and Student networks during training.

**Table 1: The statistics of datasets**

| Dataset | Nodes | Edges | Edge types | Anamaly | Features |
|---|---|---|---|---|---|
| T-Finance | 39357 | 21222543 | 1 | 4.58% | 10 |
| Tolokers | 11758 | 519000 | 1 | 21.8% | 10 |
| Amazon | 11944 | 4398392 | 3 | 6.87% | 25 |
| Elliptic | 203769 | 234355 | 1 | 9.8% | 166 |

## 4.2 Parameter Details

In this study, the training set ratio is 40% and 1% in the supervised scenario and semi-supervised scenario, while the ratio of validation and test sets is 1:2. During training, the hyperparameters for our method are configured as follows: the number of multi-head spectral filters is fixed at 3, the Chebyshev polynomial order K is set to 2, and the loss function weights are set to $\lambda_{contrast}=0.1$ and $\lambda_{div}=0.05$. We employ a Teacher warm-up period of 5 epochs. All methods are optimized using Adam and trained for 100 epochs with a learning rate of 0.01. To ensure statistical reliability, each method is executed for 10 independent runs; we report the trimmed mean of the performance metrics after excluding the highest and lowest values. The hidden layer dimension is set to 64 across all methods. All experiments were conducted on a machine equipped with 15 vCPUs (Intel Xeon Platinum 8358P @ 2.60 GHz), 90 GB RAM, and one NVIDIA RTX 3090 GPU.

The baselines compared in this study fall into two categories: traditional GNN approaches represented by GCN, GAT, GIN, and ChebyNet, and current SOTA methods represented by CARE-GNN, PC-GNN, BWGNN, and DSGAD.

## 4.3 Ablation Analysis

**Effectiveness of the Spectral-Adaptive Backbone**

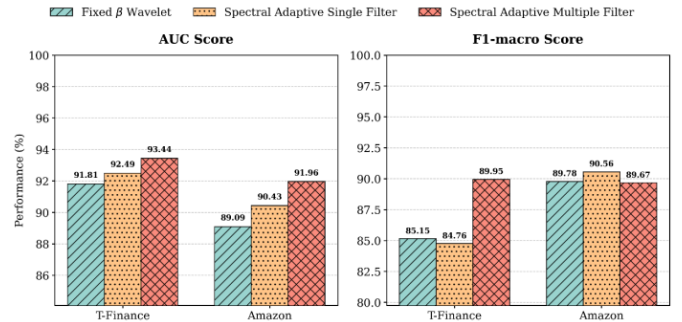

Figure 4: Analysis of the effectiveness of the spectral-adaptive backbone

To validate the performance of the spectral-adaptive multi-head filters, we compare three configurations: Fixed β Wavelet, Spectral-Adaptive Single Filter (SASF), and Spectral-Adaptive Multi-Filter (SAMF).

As shown in Figure 4, SASF achieves an AUC improvement of 0.68% on T-Finance and 1.34% on Amazon compared to Fixed β Wavelet. This is because using fixed β wavelets is equivalent to treating all subgraphs with a uniform processing strategy; while capable of covering low/mid/high frequencies, it cannot precisely match the spectral energy distribution of specific graph instances via parameter tuning. Conversely, SASF can instantly generate Chebyshev filter parameters that best fit the current graph's spectral characteristics based on the input graph structure. This instance-level adaptation enables the model to capture subtle anomaly signals that are missed by fixed filters. Additionally, we observe that the F1-macro performance of SASF on T-finance is lower than that of the Fixed β Wavelet, verifying that a single adaptive filter is insufficient to simultaneously account for features across different frequency bands.

Furthermore, compared to SASF, SAMF demonstrates superior feature decoupling capabilities. On T-finance, SAMF achieves a substantial increase of 4.8% in F1-macro; on Amazon, the AUC further improves by 1.53%. This is primarily attributed to the mixed-mode nature of financial fraud—sometimes manifesting as high-frequency local anomalies and other times as low-frequency global collusion (e.g., money laundering patterns). Our designed spectral multi-head mechanism allows different filter heads to specialize in features of specific frequency bands, avoiding the dilemma of "trading off" between different bands that limits F1-macro performance. Essentially, the multi-head design unlocks the model's full-spectrum feature capture capability.

Regarding the slight fluctuation in F1-macro on the Amazon dataset (SASF 90.56% vs. SAMF 89.67%), this reflects a fundamental shift in detection strategy. Compared to the conservative smoothing strategy of SASF, the multi-head mechanism of SAMF unleashes a keen sensitivity to high-frequency anomaly signals. Although this "aggressive" search for

## Table 2 Comparison on Amazon and T-finance datasets

| Dataset | Amazon(1%) | | Amazon(40%) | | T-finance(1%) | | T-finance(40%) | |
|---|---|---|---|---|---|---|---|---|
| Metric | F1-macro | AUC | F1-macro | AUC | F1-macro | AUC | F1-macro | AUC |
| GCN | 67.98 | 83.81 | 69.97 | 84.73 | 55.26 | 58.63 | 70.21 | 65.32 |
| GAT | 60.55 | 74.61 | 82.17 | 88.95 | 52.11 | 52.76 | 52.61 | 73.12 |
| GIN | 68.99 | 79.74 | 69.92 | 83.51 | 58.19 | 69.67 | 66.19 | 80.47 |
| ChebyNet | 86.79 | 88.35 | 92.31 | 94.52 | 77.28 | 86.93 | 81.54 | 89.41 |
| CARE-GNN | 68.58 | 87.93 | 88.19 | 91.21 | 73.33 | 90.24 | 77.98 | 92.01 |
| PC-GNN | 79.81 | 90.22 | 90.53 | 96.31 | 61.89 | 90.26 | 62.78 | 91.14 |
| BWGNN(homo) | 90.76 | 89.32 | 92.08 | 97.86 | 89.44 | 93.59 | 90.26 | 95.84 |
| BWGNN(hetero) | 83.26 | 86.45 | 92.4 | 97.84 | 88.17 | 92.74 | 88.1 | 95.6 |
| DSGAD(homo) | 86.58 | 89.17 | 92.03 | 97.83 | 89.42 | 93.11 | 90.59 | 96.03 |
| DSGAD(hetero) | 87.19 | 89.73 | 92.14 | 97.73 | - | - | - | - |
| MHSA-GNN | **91.12** | **93.96** | **92.47** | **98.27** | **90.23** | **93.79** | **91.3** | **96.65** |

## Table 3 Comparison on Tolokers and Elliptic datasets

| Dataset | Tolokers(1%) | | Tolokers(40%) | | Elliptic(1%) | | Elliptic(40%) | |
|---|---|---|---|---|---|---|---|---|
| Metric | F1-macro | AUC | F1-macro | AUC | F1-macro | AUC | F1-macro | AUC |
| GCN | 52.33 | 65.17 | 58.48 | 70.35 | 61.53 | 77.15 | 74.59 | 85.29 |
| GAT | 51.29 | 62.09 | 56.55 | 66.87 | 56.68 | 73.44 | 68.35 | 79.18 |
| GIN | 52.81 | 65.31 | 60.37 | 71.08 | 65.27 | 81.35 | 77.82 | 88.64 |
| ChebyNet | 56.87 | 69.73 | 64.78 | 74.31 | 70.51 | 83.67 | 81.33 | 91.22 |
| CARE-GNN | 53.08 | 66.34 | 59.54 | 71.36 | 71.85 | 85.74 | 82.95 | 93.29 |
| PC-GNN | 53.63 | 67.91 | 61.77 | 72.34 | 73.08 | 86.22 | 84.12 | 94.56 |
| BWGNN(homo) | 61.88 | 71.15 | 68.06 | 80.39 | 83.88 | 90.91 | **93.15** | 97.42 |
| BWGNN(hetero) | 62.1 | 72.17 | 68.49 | 81.43 | 83.73 | 88.81 | 93.09 | 97.32 |
| DSGAD | 61.94 | 71.43 | 67.83 | 80.56 | **84.18** | 89.14 | 92.08 | 97.18 |
| MHSA-GNN | **62.14** | **73.47** | **68.84** | **82.41** | 83.78 | **91.9** | 92.63 | **97.55** |

weak signals introduce a small number of false positives in the noisy Amazon dataset (causing a slight F1 dip), but it successfully uncovers a large number of concealed fraudsters missed by SASF, driving the AUC to a new high of 91.96%. In practical anti-fraud applications, a higher AUC indicates stronger potential for mining complex samples, which is more decisive than F1 fluctuations under a single threshold.

**Validation of Dual Regularization Mechanism**

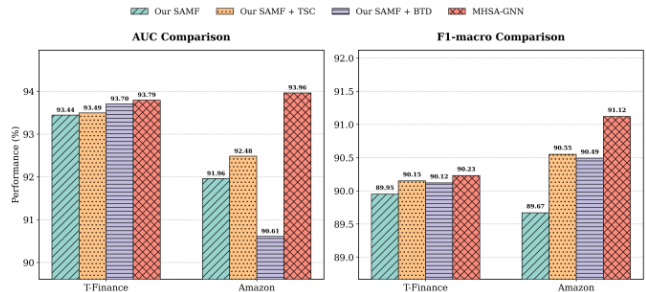

**Figure 5. Analysis of the validation of the dual regularization mechanism**

We validate the Filter Trajectory Stability Constraint (TSC) and Barlow Twins-based Decorrelation (BTD) on top of the SAMF architecture. As shown in Figure 5. Using TSC alone yields stable performance gains on both T-finance and Amazon. This proves that introducing a teacher network as an anchor guides the filters to learn meaningful features, resolving the issue of blind learning. When BTD is used in isolation, a decline in AUC is observed on the Amazon dataset. This occurs because BTD enforces orthogonality among heads; without the correct optimization direction provided by TSC, the model may "differ for the sake of differing," leading certain heads to learn orthogonal but useless noise features, thereby damaging AUC performance. This validates the necessity of our proposed dual regularization. When both are employed, performance improves dramatically, reaching 93.96% AUC on Amazon.

In summary, the TSC mechanism provides a high-quality learning target for each filter head to ensure relevance (learning "accurate" features), while BTD enforces the H high-quality filters to be distinct from one another, covering different dimensions of the feature space (learning "diverse" features). Ultimately, our model learns a set of complementary filter experts, achieving SOTA performance across multiple datasets.

### 4.4 Comprehensive Comparison

As indicated in Tables 2 and 3, GCN and GAT perform poorly on these datasets, particularly on T-finance. This is because GCN and GAT are intrinsically low-pass filters prone to smoothing neighbor features; when anomalous high-frequency signals in fraud detection are "flattened" by such smoothing, disguised fraudsters become unidentifiable. Our proposed spectral-adaptive mechanism actively preserves high-frequency signals, avoiding the over-smoothing pitfalls of classic GNNs. While ChebyNet introduces Chebyshev polynomials as band-pass filters, its coefficients are globally shared and static, lacking dynamic adaptation to subgraph instances. Methods like CARE-GNN and PC-GNN focus primarily on the spatial domain—filtering neighbors or addressing class imbalance—while neglecting the critical feature of spectral energy distribution. Regarding BWGNN, although it achieves 93.15% F1-macro on Elliptic, its performance on datasets like Tolokers is less impressive. This is because it uses fixed Beta wavelets; when the anomaly frequency distribution of a dataset (e.g., Tolokers) deviates from the preset Beta distribution, BWGNN struggles to adapt. Similarly, DSGAD, despite being a strong SOTA baseline with trainable wavelets, remains fundamentally based on linear combinations of predetermined Beta bases.

Our algorithm leverages spectral fingerprints generated from Rayleigh quotients and eigenvalue statistics to achieve instance-level parameter generation rather than simple linear combinations, effectively creating "tailor-made" adaptive filters for each graph. Furthermore, our TSC+BTD fusion scheme is logically superior to the convolutional fusion of DSGAD: TSC ensures the heads "learn accurately," while BTD ensures they "learn broadly".

Consequently, our algorithm achieves superior results across these four diverse datasets (leading in AUC across the board). Notably, on the Tolokers dataset, it achieves an AUC of 73.47% when the training ratio is 1%, significantly outperforming BWGNN and DSGAD, proving its robustness in high-heterogeneity, high-noise environments. While our model's F1-macro on the Elliptic dataset is slightly lower than the SOTA, this is acceptable because F1 scores are highly sensitive to classification thresholds. In complex anti-money laundering scenarios like Elliptic, AUC is often more critical than F1-macro as it represents the model's fundamental risk discrimination capability. Through the above comparisons and discussions, the superior performance of our model in terms of AUC and F1-macro demonstrates its advanced capabilities in learning feature representations.

## 5 Related Work

Graph Neural Networks for Anomaly Detection GNNs have shown great promise in fraud detection by aggregating neighborhood information[2-4]. Early spatial methods like GCN[5] and GraphSAGE[6] operate under the homophily assumption, effectively acting as low-pass filters[7]. However, this assumption is often violated in financial networks where fraudsters hide among normal users [1]. To address the resulting over-smoothing and heterophily issues [5, 8, 9], researchers have proposed structural learning and pruning strategies. GHRN [22] prunes inter-class edges based on high-frequency indicators, while LH-GNN [23] adjusts filtering based on estimated homophily ratios.

Spectral Graph Neural Networks Spectral GNNs filter graph signals based on the eigenvalues of the Laplacian matrix [10, 11, 13]. Polynomial Approximations: Methods like ChebNet [15], GIN [14], and DeepWalk [16] use polynomials to approximate filters. While theoretically powerful, high-order polynomials are difficult to tune and may lead to instability [20]. BernNet [12] and AutoGCN [21] improve flexibility by learning arbitrary frequency responses. To achieve better localization, GWNN utilizes Heat Kernels, which are inherently low-pass. To capture anomalies, BWGNN [24] introduces Beta distributions to model band-pass and high-pass filters, and DSGAD [26] further extends this with trainable Beta-mixture wavelets. Other works like GraphWave [24] and AGFL [25] explore unsupervised structural roles and entropy-based frequency selection, respectively.

Adaptive and Multi-Frequency Filtering Recent advancements focus on adaptive and multi-scale filtering to capture complex fraud patterns [2, 18, 19]. NMFA [29], GraphPN [30], and AdaGNN [34] employ multi-head or multi-channel mechanisms to extract features from different frequency bands. SComGNN [32] and ChiGAD [31] specifically design low/mid-pass filters for recommendation or heterogeneous graphs. GRASPED [33] utilizes Wiener filters for unsupervised reconstruction, while Balcilar et al. [2] and Bo et al. [19] analyze the necessity of explicit high-pass filtering capability.

## 6 Conclusion and Future Work

This paper proposes a novel dynamic spectral-adaptive framework, MHSA-GNN, aiming to address the lack of flexibility and over-smoothing issues inherent in traditional filters for graph anomaly detection. Unlike low-pass models such as GCN and GAT that tend to smooth high-frequency signals, our method implements instance-level generation of filter parameters via a hypernetwork and spectral fingerprint mechanism. This mechanism endows the model with the ability to actively preserve high-frequency signals, enabling effective identification of camouflaged fraudsters.

Compared to methods based on preset bases (e.g., Beta distributions) like BWGNN and DSGAD, our algorithm is no longer constrained by fixed prior distributions and can flexibly adapt to datasets with complex frequency distributions, such as Tolokers, achieving significant performance gains in AUC. Furthermore, our proposed TSC+BTD dual regularization scheme is logically superior to traditional linear convolutional fusion, effectively balancing the accuracy and diversity of the multi-head filters. Although the F1-macro score fluctuates slightly due to threshold sensitivity in specific scenarios (e.g., Elliptic), the robust performance of our model in terms of AUC demonstrates its essential superiority in risk ranking and discriminative capability. Future work will explore extending this spectral-adaptive mechanism to larger-scale dynamic graph detection tasks.

## ACKNOWLEDGMENTS

This work was supported by the Fundamental Research Funds for the Central Universities(Nos. 2025bsky023).

## REFERENCESW

[1] Pang G, Shen C, Cao L, et al. Deep learning for anomaly detection: A review. *ACM computing surveys (CSUR), 2021, 54(2): 1-38.*
[2] Balcilar M, Renton G, Héroux P, et al. Analyzing the expressive power of graph neural networks in a spectral perspective. *International Conference on learning representations. 2021.*
[3] Ma X, Wu J, Xue S, et al. A comprehensive survey on graph anomaly detection with deep learning. *IEEE transactions on knowledge and data engineering, 2021, 35(12): 12012-12038.*
[4] Rayana S, Akoglu L. Collective opinion spam detection: Bridging review networks and metadata. *Proceedings of the 21st ACM SIGKDD International Conference on Knowledge Discovery and Data Mining. 2015: 985-994.*
[5] Kipf T N. Semi-supervised classification with graph convolutional networks. *arXiv preprint arXiv:1609.02907, 2016.*
[6] Hamilton W, Ying Z, Leskovec J. Inductive representation learning on large graphs. *Advances in neural information processing systems, 2017, 30.*
[7] Zheng X, Wang Y, Liu Y, et al. Graph neural networks for graphs with heterophily: A survey. *arXiv preprint arXiv:2202.07082, 2022.*
[8] Huang W, Rong Y, Xu T, et al. Tackling over-smoothing for general graph convolutional networks. *arXiv preprint arXiv:2008.09864, 2020.*
[9] Oono K, Suzuki T. Graph neural networks exponentially lose expressive power for node classification. *arXiv preprint arXiv:1905.10947, 2019.*
[10] Bruna J, Zaremba W, Szlam A, et al. Spectral networks and locally connected networks on graphs. *arXiv preprint arXiv:1312.6203, 2013.*
[11] Tang J, Li J, Gao Z, et al. Rethinking graph neural networks for anomaly detection. *International conference on machine learning. PMLR, 2022: 21076-21089.*
[12] He M, Wei Z, Xu H. Bernnet: Learning arbitrary graph spectral filters via bernstein approximation. *Advances in neural information processing systems, 2021, 34: 14239-14251.*
[13] Chen Z, Chen F, Zhang L, et al. Bridging the gap between spatial and spectral domains: A unified framework for graph neural networks. *ACM Computing Surveys, 2023, 56(5): 1-42.*

[14] Xu K, Hu W, Leskovec J, et al. How powerful are graph neural networks?. *arXiv preprint arXiv:1810.00826, 2018.*
[15] Defferrard M, Bresson X, Vandergheynst P. Convolutional neural networks on graphs with fast localized spectral filtering. *Advances in neural information processing systems, 2016, 29.*
[16] Perozzi B, Al-Rfou R, Skiena S. Deepwalk: Online learning of social representations. *Proceedings of the 20th ACM SIGKDD international conference on Knowledge discovery and data mining. 2014: 701-710.*
[17] Veličković P, Cucurull G, Casanova A, et al. Graph attention networks. *arXiv preprint arXiv:1710.10903, 2017.*
[18] Levie R, Monti F, Bresson X, et al. Cayleynets: Graph convolutional neural networks with complex rational spectral filters. *IEEE Transactions on Signal Processing, 2018, 67(1): 97-109.*
[19] Bo D, Wang X, Liu Y, et al. A survey on spectral graph neural networks. *arXiv preprint arXiv:2302.05631, 2023.*
[20] Xu B, Shen H, Cao Q, et al. Graph wavelet neural network. *arXiv preprint arXiv:1904.07785, 2019.*
[21] Wu Z, Pan S, Long G, et al. Beyond low-pass filtering: Graph convolutional networks with automatic filtering. *IEEE Transactions on Knowledge and Data Engineering, 2022, 35(7): 6687-6697.*
[22] Gao Y, Wang X, He X, et al. Addressing heterophily in graph anomaly detection: A perspective of graph spectrum. *Proceedings of the ACM Web Conference 2023. 2023: 1528-1538.*
[23] Wo Z, Shao M, Zhang S, et al. Local Homophily-Aware Graph Neural Network with Adaptive Polynomial Filters for Scalable Graph Anomaly Detection. *Proceedings of the 31st ACM SIGKDD Conference on Knowledge Discovery and Data Mining V. 2. 2025: 3180-3191.*
[24] Donnat C, Zitnik M, Hallac D, et al. Learning structural node embeddings via diffusion wavelets. *Proceedings of the 24th ACM SIGKDD international conference on knowledge discovery & data mining. 2018: 1320-1329.*
[25] Tu B, Yang X, He B, et al. Anomaly detection in hyperspectral images using adaptive graph frequency location. *IEEE Transactions on Neural Networks and Learning Systems, 2024.*
[26] Zheng J, Yang C, Zhang T, et al. Dynamic Spectral Graph Anomaly Detection. *Proceedings of the AAAI Conference on Artificial Intelligence. 2025, 39(12): 13410-13418.*
[27] Li Z, Liu R, Chen D, et al. OR-Gate Mixup Multiscale Spectral Graph Neural Network for Node Anomaly Detection. *IEEE Transactions on Neural Networks and Learning Systems, 2025.*
[28] L. Tan, R. Dian, S. Li and J. Liu, "Frequency-Spatial Domain Feature Fusion for Spectral Super-Resolution," in *IEEE Transactions on Computational Imaging, vol. 10, pp. 589-599, 2024.*
[29] Xu P, Wang Y, Wu M, et al. NMFA: Node-Level Multi-Frequency Adaptive Graph Neural Network for Graph Fraud Detection[C]//*Proceedings of the 2024 4th International Conference on Artificial Intelligence, Big Data and Algorithms. 2024: 1094-1100.*
[30] Zhang J, Yu L, Huang Z, et al. Topology augmented multi-band and multi-scale filtering for graph anomaly detection. *ACM Transactions on Knowledge Discovery from Data, 2025, 19(8): 1-27.*
[31] Li X, Dong X, Zhang X, et al. Chi-Square Wavelet Graph Neural Networks for Heterogeneous Graph Anomaly Detection. *Proceedings of the 31st ACM SIGKDD Conference on Knowledge Discovery and Data Mining V. 2. 2025: 1565-1576.*
[32] Luo H, Meng X, Wang S, et al. Spectral-based graph neural networks for complementary item recommendation. *Proceedings of the AAAI Conference on Artificial Intelligence. 2024, 38(8): 8868-8876.*
[33] Choong W H, Liu J, Kao C Y, et al. GRASPED: Graph Anomaly Detection using Autoencoder with Spectral Encoder and Decoder (Full Version). *arXiv preprint arXiv:2508.15633, 2025.*
[34] Dong Y, Ding K, Jalaian B, et al. Adagnn: Graph neural networks with adaptive frequency response filter. *Proceedings of the 30th ACM international conference on information & knowledge management. 2021: 392-401.*
[35] Chen T, Trogdon T, Ubaru S. Analysis of stochastic Lanczos quadrature for spectrum approximation. *International Conference on Machine Learning. PMLR, 2021: 1728-1739.*
[36] Tarvainen A, Valpola H. Mean teachers are better role models: Weight-averaged consistency targets improve semi-supervised deep learning results. *Advances in neural information processing systems, 2017, 30.*
[37] Zbontar J, Jing L, Misra I, et al. Barlow twins: Self-supervised learning via redundancy reduction. *International conference on machine learning. PMLR, 2021: 12310-12320.*
[38] Julian John McAuley and Jure Leskovec. 2013. From amateurs to connoisseurs: modeling the evolution of user expertise through online reviews. *In Proceedings of the 22nd International Conference on World Wide Web (WWW '13). Association for Computing Machinery, New York, NY, USA, 897–908.*
[39] Likhobaba D, Pavlichenko N, Ustalov D. Toloker graph: Interaction of crowd annotators[EB/OL].(2023)
[40] Weber M, Domeniconi G, Chen J, et al. Anti-money laundering in bitcoin: Experimenting with graph convolutional networks for financial forensics. *arXiv preprint arXiv:1908.02591, 2019.*
[41] You Y, Chen T, Sui Y, et al. Graph contrastive learning with augmentations. *Advances in neural information processing systems, 2020, 33: 5812-5823.*